\documentclass[conference]{IEEEtran}
\IEEEoverridecommandlockouts
\usepackage{cite}
\usepackage{amsmath,amssymb,amsfonts}
\usepackage{algorithmic}
\usepackage[ruled,vlined]{algorithm2e}
\usepackage{graphicx}
\usepackage{textcomp}
\usepackage{xcolor}
\usepackage{multirow}
\usepackage{verbatim}
\usepackage{longtable}
\def\BibTeX{{\rm B\kern-.05em{\sc i\kern-.025em b}\kern-.08em
    T\kern-.1667em\lower.7ex\hbox{E}\kern-.125emX}}
\begin{document}

\title{\vspace{6mm}Multi-Object Grasping -- Types and Taxonomy 
}
\author{Yu Sun, Eliza Amatova, and Tianze Chen
\thanks{The authors are from the Robot Perception and Action Lab (RPAL) of Computer Science and Engineering Department, University of South Florida, Tampa, FL 33620, USA. Email: \texttt{\{yusun,amatovae,tianzechen\}@usf.edu}. Eliza Amatova is an undergraduate student.}
}

\maketitle

\begin{abstract}
This paper proposes 12 multi-object grasps (MOGs) types from a human and robot grasping data set. The grasp types are then analyzed and organized into a MOG taxonomy. This paper first presents three MOG data collection setups: a human finger tracking setup for multi-object grasping demonstrations, a real system with Barretthand, UR5e arm, and a MOG algorithm, a simulation system with the same settings as the real system. Then the paper describes a novel stochastic grasping routine designed based on a biased random walk to explore the robotic hand's configuration space for feasible MOGs. Based on observations in both the human demonstrations and robotic MOG solutions, this paper proposes 12 MOG types in two groups: shape-based types and function-based types. The new MOG types are compared using six characteristics and then compiled into a taxonomy. This paper then introduces the observed MOG type combinations and shows examples of 16 different combinations. 
\end{abstract}

\section{Introduction}

Grasping multiple objects at once from a pile is common for us. It is so common that we have the word ``handful'' to describe a quantity that fills the hand. When we were children, we grasped a handful of candies from a bowl. We pick a handful of Brussel sprouts from a bag when we cook. When we make a drink, we pick up two or three ice-cube at once from an ice bucket. If we want multiple objects, we pick them up rarely one by one, but at once, because it is more efficient. We can observe similar scenarios in manufacturing and logistics. In manufacturing, workers get a handful of bolts from a bin and then put them on one by one. In logistics, warehouses regularly apply a strategy called batch picking and instruct pickers to collect the same items for multiple orders simultaneously. Human workers are very good at picking multiple same items at once. 

However, robotic technology today has only been developed to pick up one thing at a time \cite{agrawal2010vision}. No matter how efficient a single-object picking is, it cannot compete with a system that can pick up multiple objects in one grasp. For example, if a robotic system can pick up an object and drop it into a bin in $3$ seconds, to pick up five same items, the robot would need to pick five times, which is $15$ seconds. In contrast, a human worker could pick five items at once in about $3$ seconds. It is five times faster. There is no way speeding up the robot could catch the difference. 

Therefore, a robot needs to gain multi-object grasping capability (MOG). However, picking up multiple objects has not been studied in robotics literature since it is widely considered difficult \cite{sun2021research}. To develop suitable tools and technologies for multi-object gasping (MOG), we should analyze the unique characteristics of multi-object grasping.

In the traditional single-object grasping (SOG), grasps are divided into precision grasps and power grasps \cite{Napier1956}. A power grasp is chosen for stability and security since it provides large contact areas between the surfaces of a hand's finger and palms and the grasped object. A precision grasp is chosen for dexterity and sensitivity since only the fingertips contact the object. When grasping multiple objects, the hand needs to provide multiple contact areas that support and press on the objects and enough space so that an adequate number of objects are in the grasp. The fingers cannot squeeze on the objects as hard as the power grasps since objects could be squeezed out.

This paper first designs three data collection approaches to discover distinctive grasp types for multi-object grasping. The first one collects multi-object grasping data performed by a human demonstrator because human grasp strategies have inspired the development of many robotic grasping approaches \cite{lin2012learning, lin2014grasp, lin2015robot}. The other two collect multi-object grasping in a simulation and a real robotic system through a stochastic grasping approach. Then, we manually identify distinctive grasp types based on the MOG data in all demonstrations. In the end, we compile all grasp types into a comprehensive MOG taxonomy.

\subsection{Related works}
The literature provides a wide range of single object grasping types and taxonomies. One of the studies classified hand usage by monitoring every action of subjects during a typical day and the authors came up with four features that annotate grasps: hand shape, force type, direction, and flow \cite{liu2014taxonomy}. Another study that was performed by Cutkosky \cite{cutkosky1989grasp} monitored the grasps used by machinists and developed a taxonomy based on the observations. The developed taxonomy consists of a hierarchical tree of grasps that begins with the two basic categories: power grasp and precision grasp. We get a more detailed task and object geometry down the tree. 
Fiex et al. \cite{feix2015grasp} developed a grasp taxonomy by comparing all other human grasp taxonomies and finding the largest set of distinct grasps for holding one object securely. Their taxonomy consists of 33 different grasp types that are arranged according to the opposition type, virtual finger assignments, type in terms of power/precision/intermediate grasp, and the thumb position. 

Most of the existing grasping taxonomies divide grasp types into power, precision, and intermediate grasps. A power grip is a grip formed with partly flexed fingers and the palm, while in precision grip, the object is pinched between fingers \cite{liu2014taxonomy}. The intermediate grasp is a mix of power and precision grasps. The existing single-object grasping taxonomies also classify the grasp types according to the position of the thumb. The thumb can be abducted, adducted, extended, or flexed. The opposition type is another criterion according to which the grasp types can be arranged. The three basic opposition types are pad opposition which occurs along a direction parallel to the palm; palm opposition which occurs along a direction perpendicular to the palm; and side opposition which occurs along a direction transverse to the palm \cite{feix2015grasp}.
In some cases, several fingers apply forces in the same direction and act in unison. Therefore, they work together as a single virtual finger (VF) \cite{feix2015grasp}. Grasp types can be classified according to the virtual finger assignments. 

Our previous work \cite{dai2013functional, lin2013grasp} has used functional principal component analysis (fPCA) and trajectory distances to analyze grasping motions and extract dynamic features in finger motions. After analyzing and comparing many finger motions of the 15 grasp types in the Cutkosky grasp taxonomy, we found several grasp types are very similar in terms of the finger motions. Based on the finger motion alone, the 15 grasp types can be grouped to 5 distinctive groups, and the finger motions from different grasp types within each group are indistinguishable.  

A limited amount of work on grasping multiple objects has been carried out for static grasp stability analysis. \cite{harada1998enveloping, harada2000rolling} discuss the enveloping grasp of multiple objects under rolling contacts and  \cite{harada2002active} studied force closure of multiple objects. It builds the theoretical basis for later work on active force closure analysis for the manipulation of multiple objects in \cite{harada2002active}. 
\cite{yoshikawa2001optimization, yamada2005grasp, yamada2015static} try to achieve stably grasping of multiple objects through force-closure-based strategies.  Our recent work \cite{chen2021multie} has studied the tactile sensing aspect of MOG and developed a deep learning approach to estimate the number of objects in the grasp. We have also developed an Markov-decision-process-based MDP-MOG model to generate an optimal policy for picking and transferring multiple same objects \cite{shenoy2021multiobject} from one bin to another. 

\section{Data collection}

\begin{figure*}[t!]
    \centering
    \includegraphics[height=0.24\linewidth]{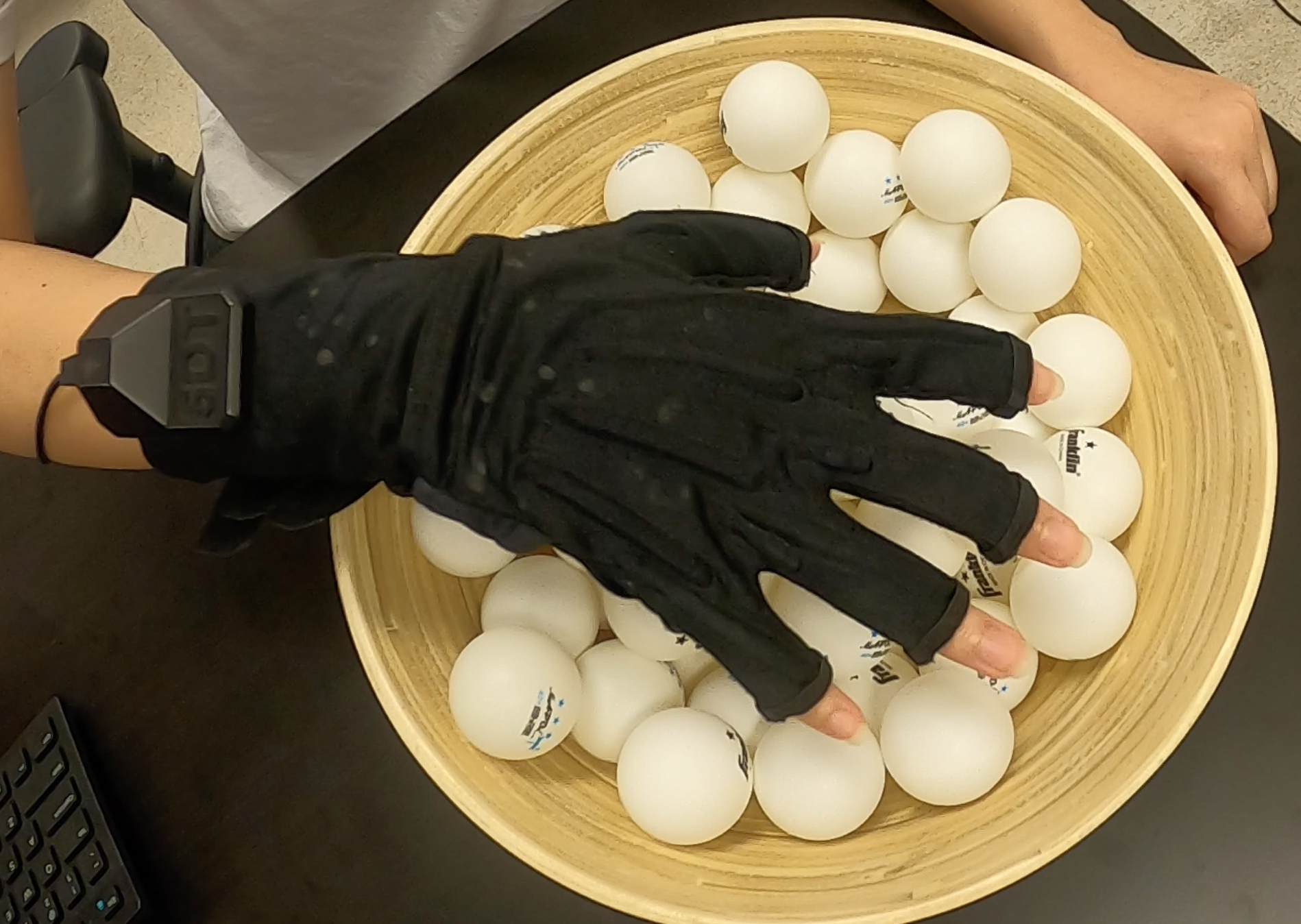}
        \includegraphics[height=0.24\linewidth]{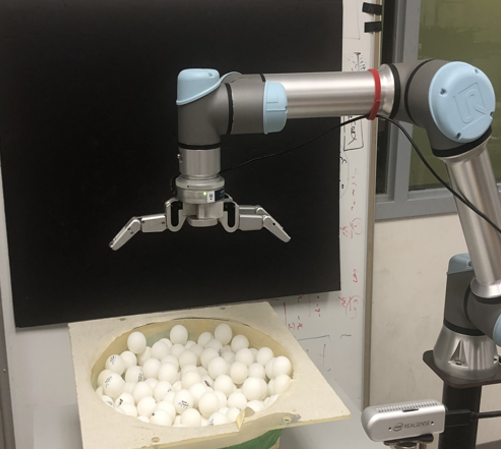}
        \includegraphics[height=0.24\linewidth]{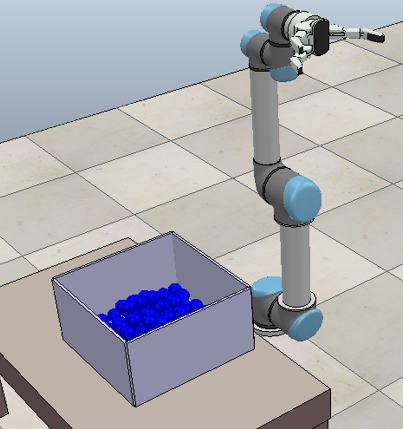}\\
    ~~~~~~~(A)~~~~~~~~~~~~~~~~~~~~~~~~~~~~~~~~~~(B)~~~~~~~~~~~~~~~~~~~~~~~~~~~~~(C)
    \caption{Data collection setups for (A) human demonstration; (B)real robotic grasping; (C) simulation environment developed in CoppeliaSim.  }
    \label{fig-system_setup}
\end{figure*}

To fully discover the multi-object grasp types and study all types of interactions between fingers and objects, we have collected multi-object grasping samples from human demonstration trials, random trials (through stochastic grasping) in a simulation setup with a simulated robotic hand and arm, and refined random trials in a real grasping setup with a robotic hand and arm.  The three setups are shown in Figure \ref{fig-system_setup}. 

\paragraph{Human demonstration setup}
As shown in Figure \ref{fig-system_setup}(A), we have collected hand posture data through a 5-DT Data Glove and video recording of a set of multi-object grasping by a human participant from a container containing about 30-50 objects. The 5-DT Data Glove has 14 sensors that measure the flexion of Metacarpophalangeal Joints and Proximal Interphalangeal Joints and the abductions between the fingers.  We use the data glove data to observe the occluded fingers when there are occlusions in the recordings. More details on the setup can be found at \cite{huang2019dataset}.

\paragraph{Real system setup}

We utilize a UR5e robot arm and a Barrett hand with tactile sensors in the PI's lab to set up a real environment and collect multi-object grasping data. The Barrett hand has seven joints, and each hand pose contains the readings from all the $7$ joints.  
The Barrett hand's palm has $24$ tactile sensors, while each finger has $24$ tactile sensors.  Each finger also has a strain gauge sensor measuring the coupled joint torque.   

\paragraph{Simulation setup}
Simulation provides an economical way of collecting a significant amount of grasping data. Figure \ref{fig-system_setup}(B) shows a preliminary setup we have developed in CoppeliaSim \footnote{The CoppeliaSim software is from: https://www.coppeliarobotics.com}. The current simulation system, as shown in Figure \ref{fig-system_setup} (C) has a UR5 robotic arm and a Barrett Hand embedded with tactile sensors and joint torque sensors. We selected CoppeliaSim because our preliminary study showed the data collected in CoppeliaSim closely resemble the data collected in the real system. 

\begin{figure*}[h]
    \centering
    \includegraphics[width=\textwidth]{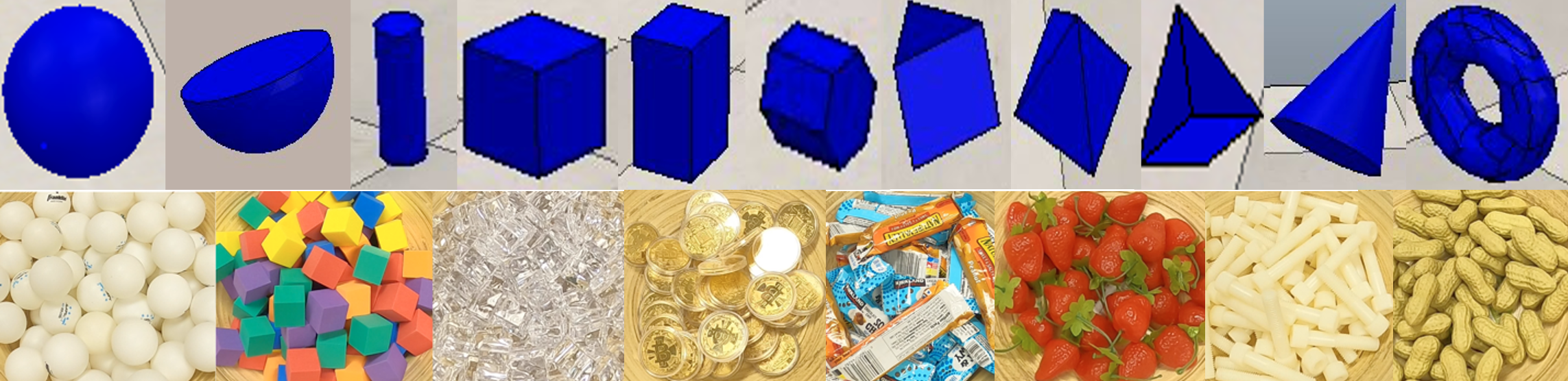}
    \caption{(Top) Basic shapes for data collection; (Bottom) Objects used in our preliminary studies.}
    \label{fig-objects}
    \vspace{-0.5cm}
\end{figure*}

\paragraph{Objects}
In the simulation, we have the robotic hand to grasp objects in eleven basic shapes: sphere, hemisphere, cuboid, cone, square pyramid, triangular pyramid, cylinder, hexagonal prism, triangular prism,  rectangular prism, and torus, and in two sizes. We have created all the proposed basic shapes in CoppeliaSim as shown in Figure \ref{fig-objects}(Top). In the human grasping demonstration and the real robotic grasping, the hands grasp eight objects with different shapes and sizes. They are ping-pong ball, foam cube, ice cube, coin, candy bar, strawberry, bolt, and peanut as shown in Figure \ref{fig-objects}(Bottom).

\subsection{Human multi-object grasping data collection}

For each shape, our demonstrator wore a data glove and grasped objects from the bowl 25 times in the following order: 5 grasps of one object, 5 of two objects, 5 of three objects, 5 of four objects, and 5 of five objects. However, the data collection routine for pencils was a little different. A total of 15 grasps were performed in the following order: 5 grasps of a minimum number (4-5 pieces) of pencils, 5 grasps of an average number of pencils, and 5 grasps of a maximum number of pencils. In each grasp, the participant immersed the hand into the container to get the required number of objects. Then the participant lifted the hand out and held it in the air to have the camera system fully observe the hand and the objects inside of the hand.  

\subsection{Robot grasping data collection}
We have developed a stochastic hand flexing/extending routine and a random stop mechanism to explore potential good hand configurations for grasping multiple objects. For each shape and size, the Barrett hand keeps performing stochastic grasps to grasp objects from the bin until a total of 100 successful cases are collected. At the end of each stochastic grasp, the hand lifts out, and we count how many objects are successfully grasped. The successful cases are the ones where the hand holds two or more objects in the air. The system records those cases with the hand pose and side photos for both systems, and object locations in hand for the simulation. 

\paragraph{Stochastic grasping routine}

The proposed stochastic flexing routine will allow the fingers to flex/extend at various rates so that a robotic hand can form different kinds of shapes. It provides the necessary diversity in exploring MOG. The stochastic flexing routine is designed based on the biased random walk idea, which is summarized in equation \ref{eq-brw}.
\begin{equation} \label{eq-brw}
d(t) = p*d_{fw} + q*d_{bw} + (1-p-q)*d_{stay},
\end{equation}
$d(t)$ represents the distance a particular finger should move at time $t$, $p$ is the probability of flexing, q is the probability of extending, $d_{fw}$ is the flexing distance, $d_{bw}$ is the extending distance, $d_{stay}$ means the finger will stay at the same location. We choose $p$ as $0.7$, $q$ as $0.1$, $d_{fw}$ as $3^{\circ}$ and $d_{bw}$ as $-3^{\circ}$.

The stochastic flexing/extending routine based on the biased random walk is detailed in \cite{chen2022multi}.
The bias random walk approach allows the fingers to flex most of the time, extend sometimes, and stop a few times. This approach simulated the human finger movement when grasping multiple objects.  We selected $3^{\circ}$/step in consideration of the time-cost of the data collection. The coupled joint of each finger moves with the base joint at a ratio of $1/3$.

\section{Multi-Object Grasp Types}

\begin{table*}[t!]
\begin{center}
\begin{tabular}{ |p{3cm}||p {2cm}|p {2cm}|p {2cm}|p {2cm}|p {2cm}|p {2cm} |}
 \hline
Types (shape-based) & Cylindrical & Funnel & Cup & Tracks & Inverse basket & Max\\ \hline
Human hand examples &
\includegraphics[width=2cm]{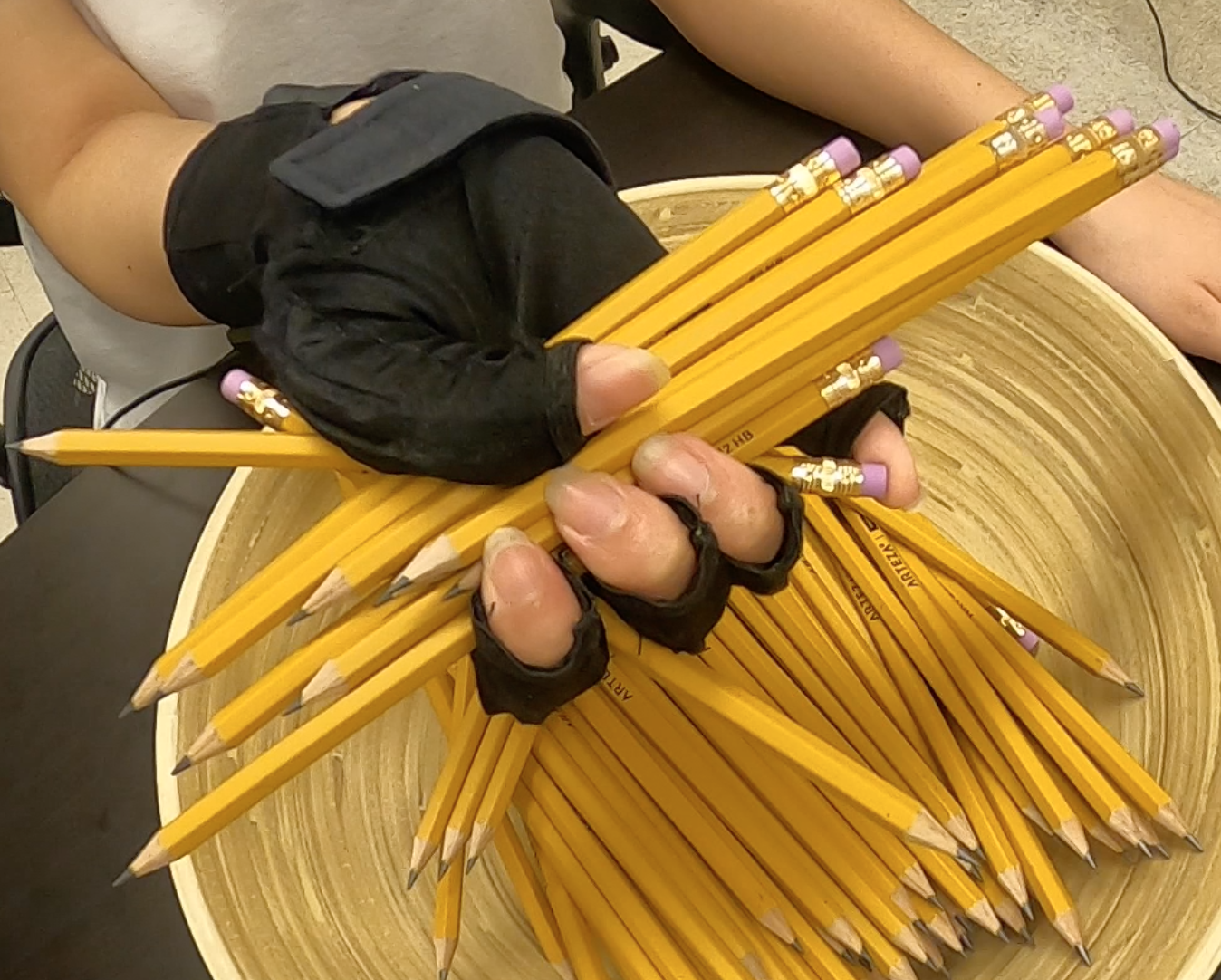} &
\includegraphics[width=2cm]{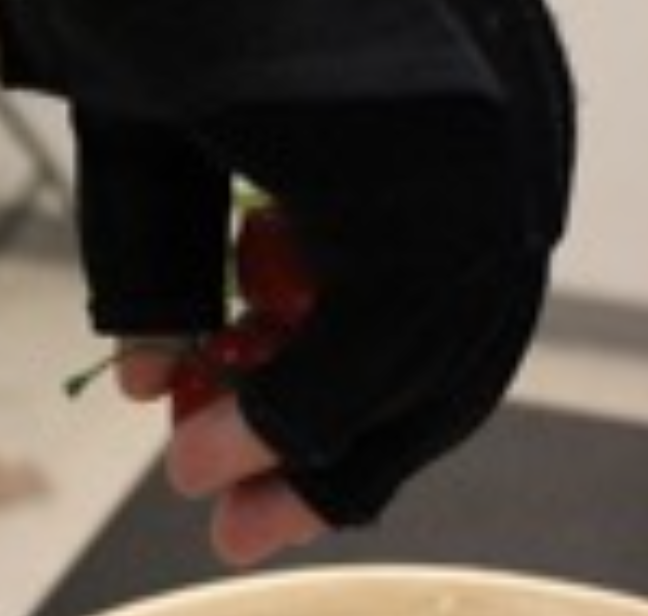} &
\includegraphics[width=2cm]{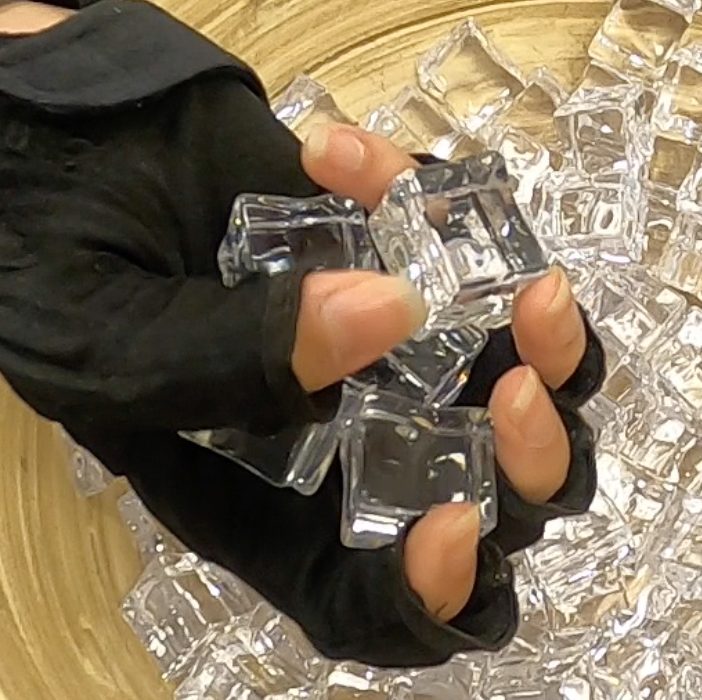} &
\includegraphics[width=2cm]{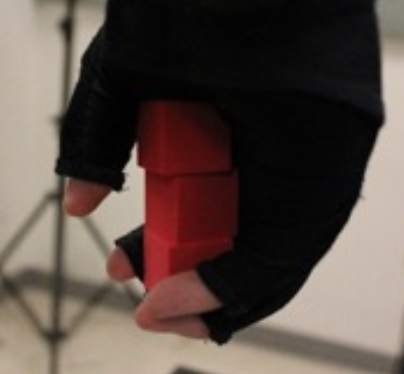} & \includegraphics[width=2cm]{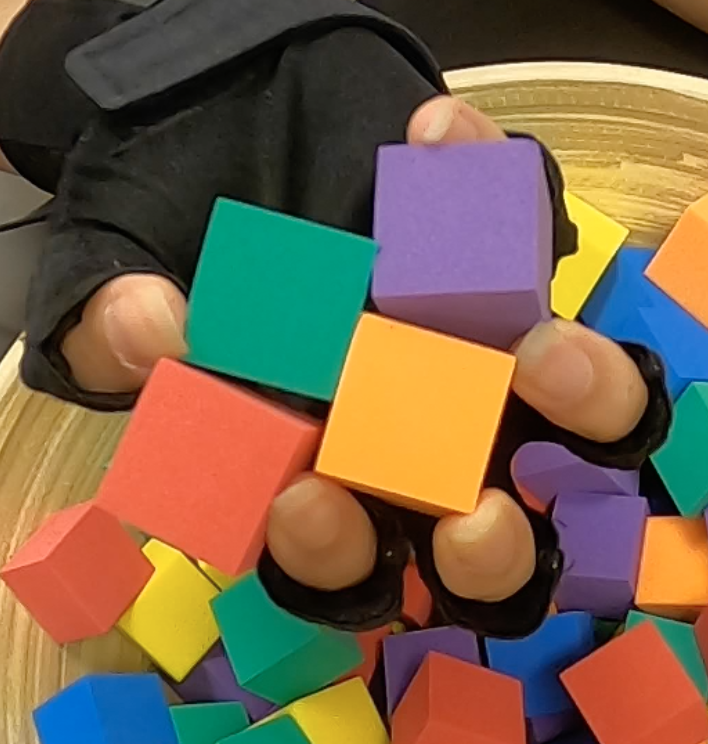} & \includegraphics[width=2cm]{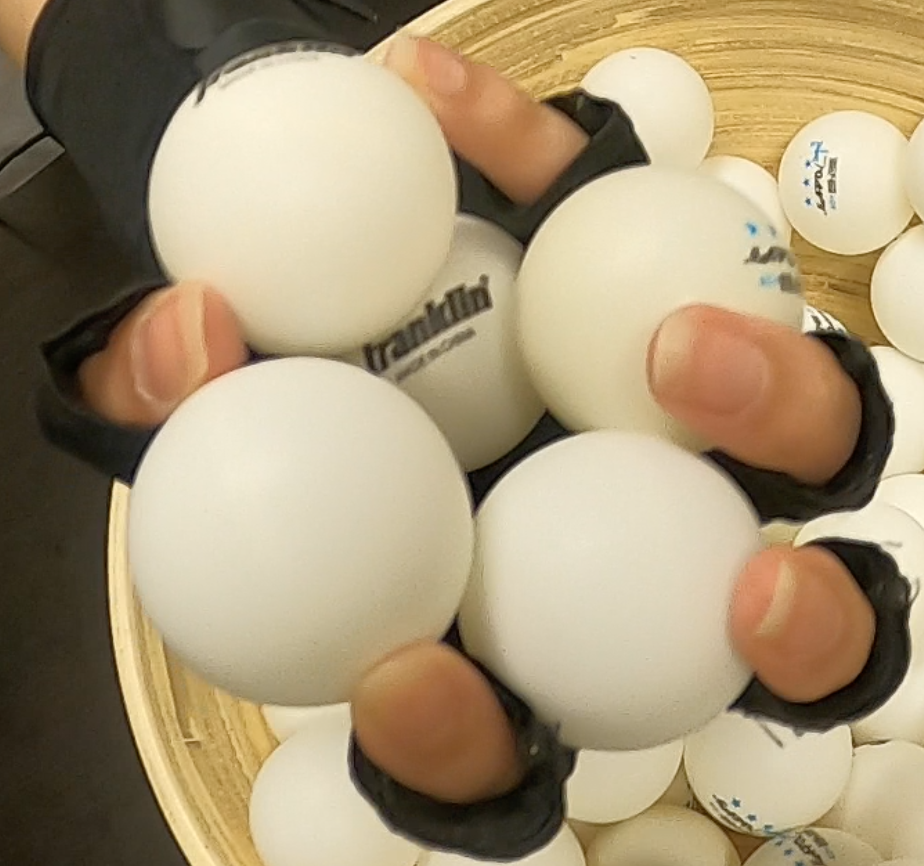} \\ 
\hline
Barrett hand examples &\includegraphics[width=2cm]{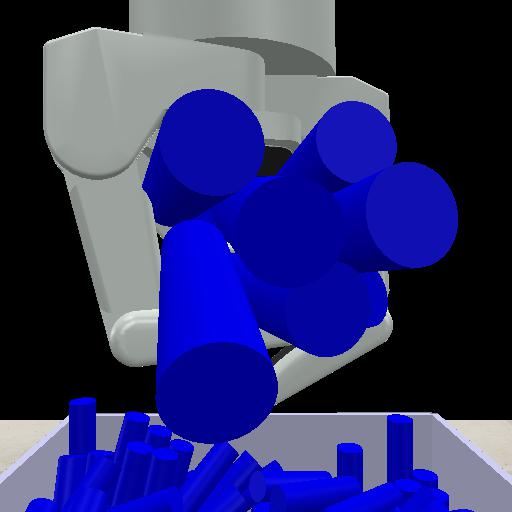} & 
\includegraphics[width=2cm]{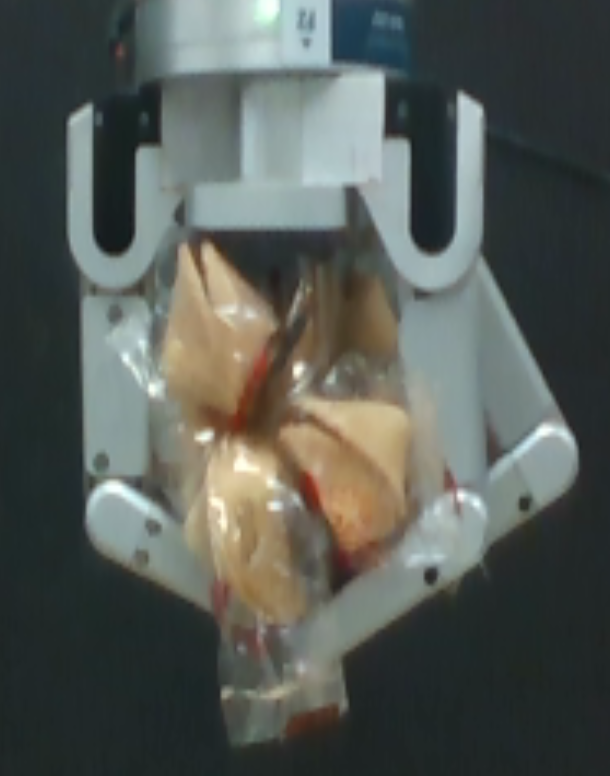}
& \includegraphics[width=2cm]{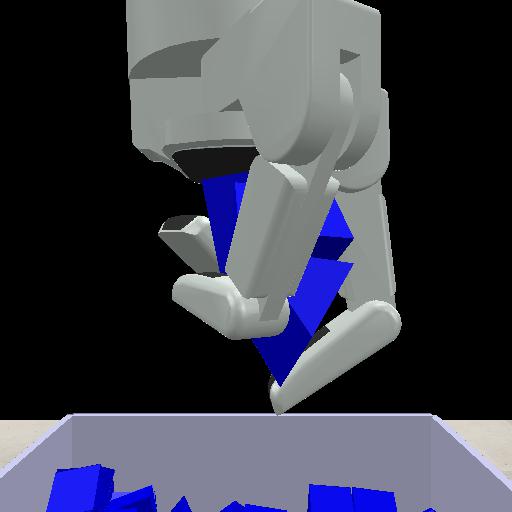} & \includegraphics[width=2cm]{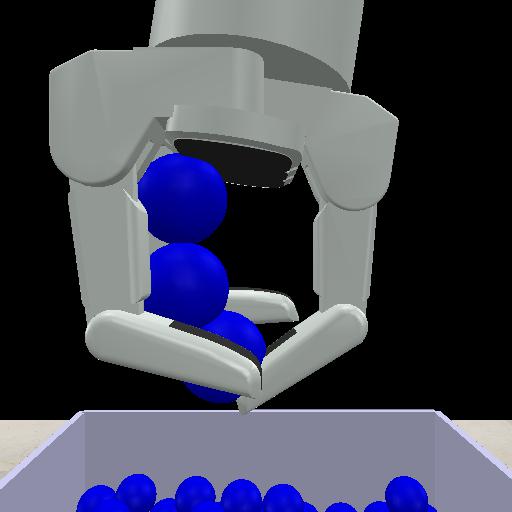}    & N/A & N/A
\\ \hline
\hline
Types (function-based) & Abduction Clip & Finger-palm clip & Finger-finger clip &  Fingertip-fingertip pinch & Fingertip-finger pinch & Multi-finger pinch\\ \hline
Human hand examples &
\includegraphics[width=2cm]{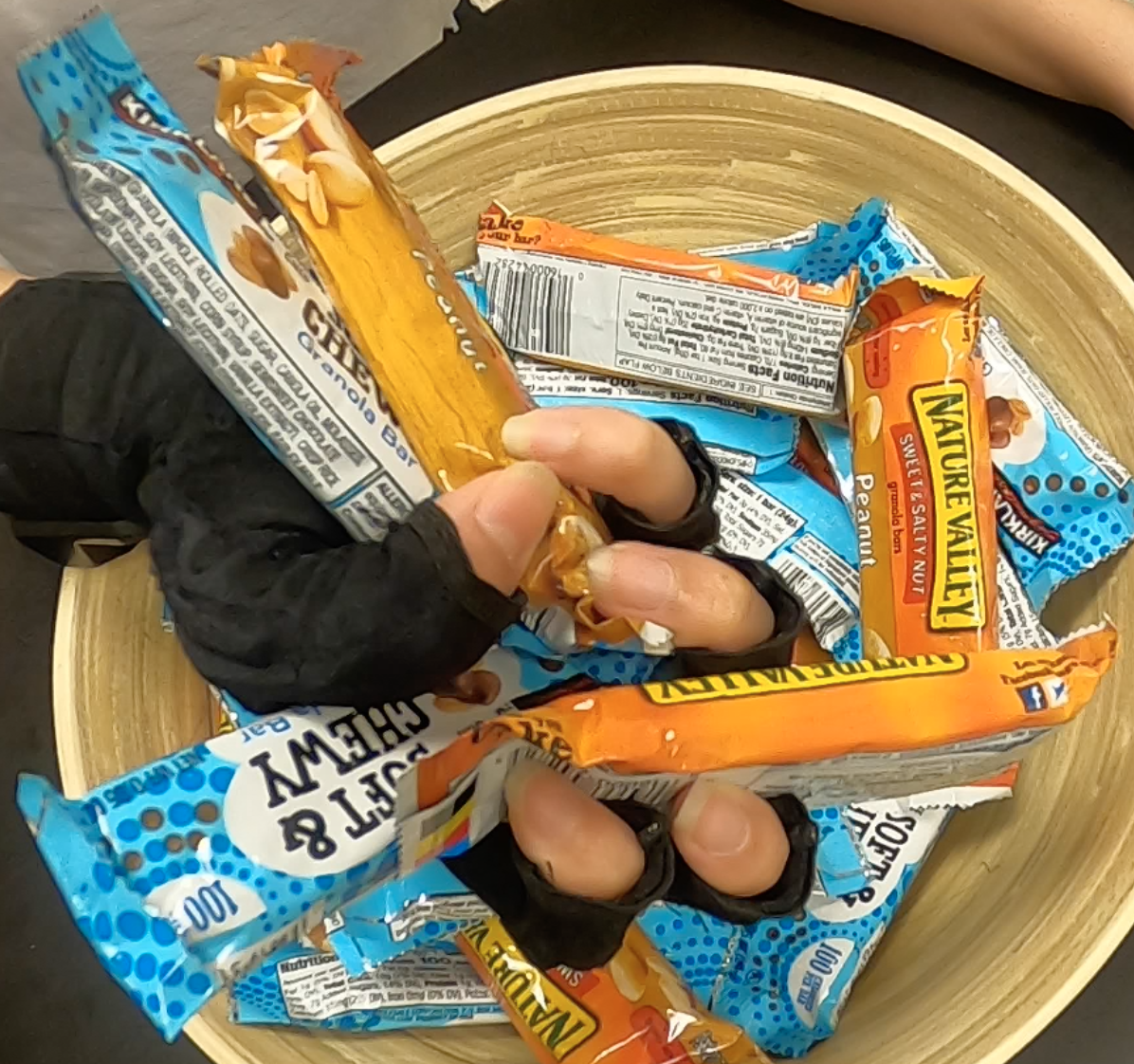} &
\includegraphics[width=2cm]{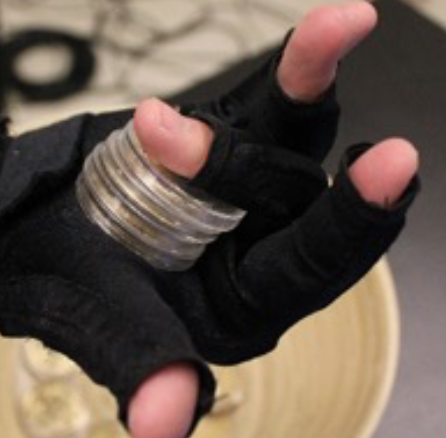} &
\includegraphics[width=2cm]{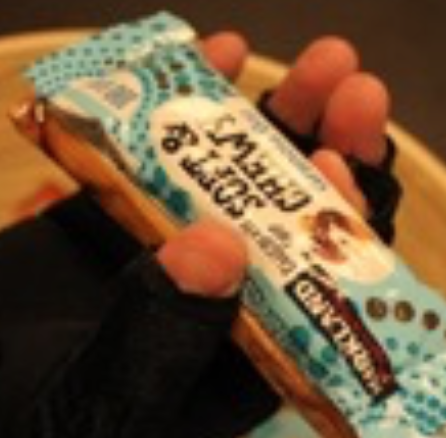} &
\includegraphics[width=2cm]{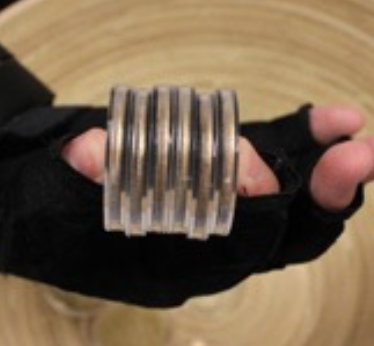} & \includegraphics[width=2cm]{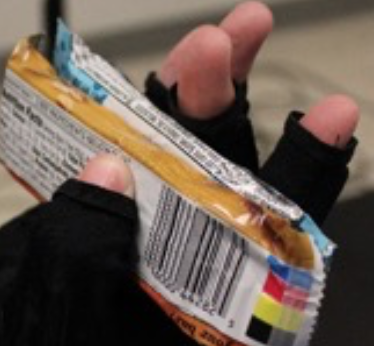} & \includegraphics[width=2cm]{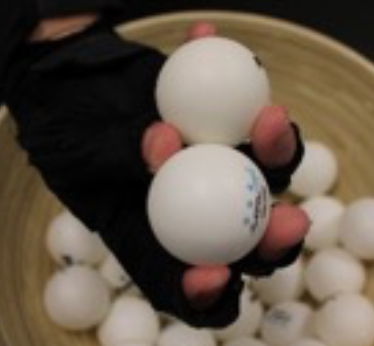} \\ 
\hline
Barrett hand examples & 
\includegraphics[width=2cm]{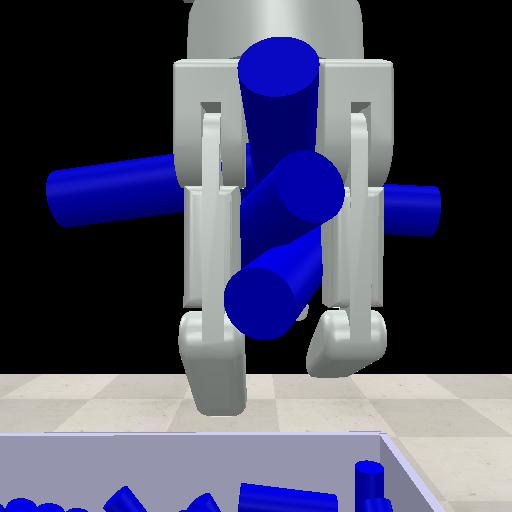} &
\includegraphics[width=2cm]{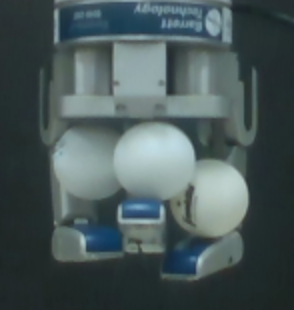} & 
N/A&
\includegraphics[width=2cm]{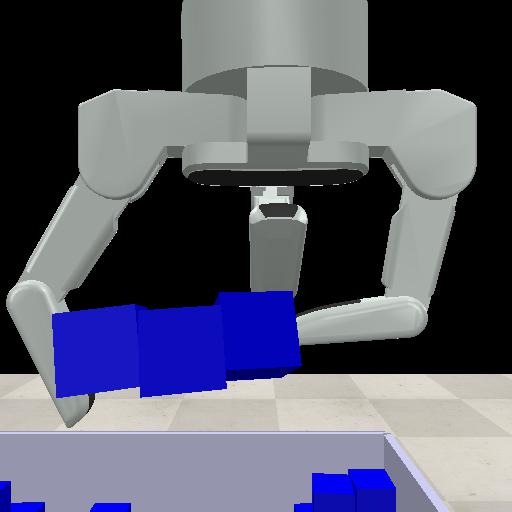} & \includegraphics[width=2cm]{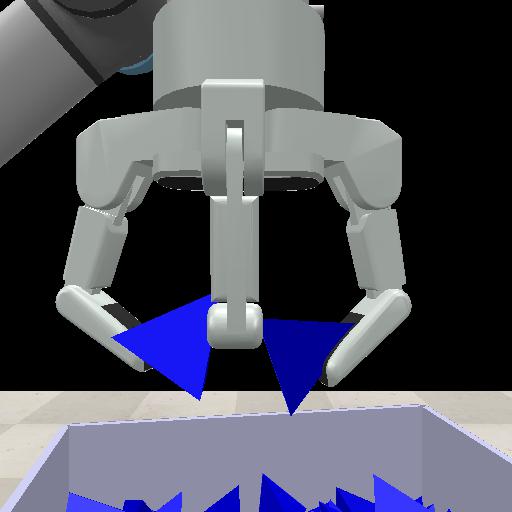}  &  N/A
\\ \hline
\end{tabular}
\caption{Grasp types.}
\vspace{-1.cm}
\label{table-types}
\end{center}
\end{table*}

\subsection{Multi-object grasp types}
Using the setups, we have collected 400 instances of human hand MOG, 26,200 instances of Barrett Hand MOG in simulation, and 1,300 instances of Barrett Hand MOG in real. We have reviewed all the instances and categorized all the instances into 12 basic types of two groups: shape-based MOG group and function-based MOG group. 

\paragraph{Shape-based MOG group} 
Both a human hand and a robot hand tend to form certain shapes when grasping multiple objects. The shapes usually involve all fingers. We have observed the following common shapes:
\begin{itemize}
    \item Cylindrical shape. The fingers and thumb form a cylindrical shape. 
    \item Funnel shape. The fingers form a funnel shape, and the fingertips form and control the funnel's opening. 
    \item Cup shape. All fingers are bending upward to form a space to hold the objects. This shape can resemble a spoon as well. 
    \item Tracks. Two fingers form two tracks so that the objects stay on the tracks, while a third finger may also be applied to block the end of the tracks to prevent the objects from falling. 
    \item Inverse basket shape. Fingers form a shape like a wire basket. Since the opening of the hand is downward when picking up objects, the shape is an inverse basket shape. Unlike the funnel shape grasp, the inverse basket shape grasp has a much larger volume and wider opening. The objects are squeezed together to prevent falling while the objects in the funnel shape grasp could rest on the fingertips. 
    \item Maximum space shape. All fingers and the thumb spread as wide as possible and then flex. They form a maximum space when they hold objects. 
\end{itemize}

\paragraph{Function-based MOG group } In some other instances, the fingers move toward each other without aiming to form a certain shape. The finger moving direction serves as a function of the finger, and the hand's final shape is less important. The function-based MOG types usually involve a subset of fingers. The types in this group are:
\begin{itemize}
\item Abduction clip (scissors). Objects are grasped in between the sides of two fingers. The two fingers act like a pair of scissors. 
\item Finger-palm clip. A finger presses the objects against the palm.  
\item Finger-finger clip. Two fingers (or one finger and one thumb) press against each other to hold objects between the fingers. It is different from the abduction clip since the finger-finger clip grasp uses the palmar side of the fingers to press on the objects. 
\item fingertip-fingertip pinch. Two fingertips use their tips pressing on the objects to hold them.  
\item fingertip-finger pinch. A fingertip uses its tip to press the objects on another finger.   
\item multi-finger pinch (tripod). Multiple fingertips use their tips pressing on the objects to hold them. They form tripod support of the objects.  
\end{itemize}

\section{Relationships of Multi-Object Grasp Types}

\subsection{Type characteristics}
We have also studied the characteristics associated with each MOG grasp type and summarized them in Table \ref{table-types}. Many types share the same characteristics. 


\paragraph{Opposable thumb/fingers} Some MOG types in Table \ref{table-types} have to have a finger at the opposite side of the other fingers, such as the Finger-finger pinch type. The finger on the opposite side can be viewed as the opposable thumb. All fingers should be on one side for some other grasp types, such as the {\it Cup} type. For the rest, having an opposable thumb is irrelevant to the type. 

\paragraph{Finger abduction/adduction} Several MOG types in Table \ref{table-types} require the fingers to abduct to an extensive or maximum range, such as the {\it Max} and {\it Finger-finger pinch} types. Some MOG types prefer to have the fingers partially abducted to form a gap between fingers, and the gap size is related to the object size, such as the {\it Track} and {\it Cup} types. In some MOG types, the fingers should be adducted, such as the {\it Cylindrical} type. In others, the abduction/adduction is irrelevant, such as the {\it Finger-palm clip} type. 

\paragraph{Finger flexion} Several MOG types in Table \ref{table-types} depend on the fingers to flex to a certain level.  Some types require the fingers to flex moderately to form a shape, such as the {\it Funnel} and {\it Track} types, while some others would need the fingers to flex quite much to form a shape, such as the {\it Cup} type. Some types would require the fingers to flex until they reach a certain resistance level. In the function-based types, fingers usually flex till reaching a certain resistance. 

\paragraph{Contact location} Different MOG types could expect different parts of the hand to be in contact with the objects.  For example, a {\it Cylindrical} grasp would expect both the fingers and the palms to contact the objects, while {\it Scissors} grasp would only expect the sides of two fingers to contact the objects.  Understanding the potential contact locations could help in designing tactile sensor locations.  

\paragraph{Manipulation difficulty} 
After a grasp is formed, the objects in the grasps of some shape-based types are not sensitive to the finger movements, such as the {\it Cup} type.  Therefore the fingers can move to manipulate the objects slowly without worrying about dropping them.  However, the grasps of the function-based grasping types could be very sensitive to finger movement and make manipulating objects difficult. In-hand manipulation strategies \cite{fearing1986implementing,rus1999hand,okamura2000overview, bicchi2000hands, ma2011dexterity} such as rolling, sliding, finger gaiting, and regrasping can be easily applied to grasps in the shape-based types but difficult to the ones in the function-based grasp types. 

\paragraph{Closure and force} Picking out objects from a bin doesn't require form closure. So, MOG grasps of any type don't have to have form closure. The grasps of the function-based grasp types usually require force-closure, such as the {\it Finger-finger pinch} type. The force closure should be calculated in considering both the finger-object and object-object contacts. The grasps of some shape-based grasp types don't even require force-closure since the objects can rest on some parts of the fingers and can rotate and move in a small range, such as the {\it Track} and {\it Cup} types.  In those grasps, the object may fall off the hand if it moves out of local containment due to shaking or other disturbances.  So they are in containment, but not in force or form closure or cage. 

\begin{table*}[t!]
\begin{center}
\begin{tabular}{ |p{2.5cm}||p {1cm}|p {1cm}||p {1cm}|p {1.3cm}|p {1.3cm}|p {1cm}|p {1.5cm}|p {1.5cm}|} 
\hline
\multirow{2}{4em}{Grasp types} & 
\multicolumn{2}{|c||}{Groups} & \multicolumn{6}{|c|}{Characteristics} \\
& Shape-based & function-based & opposable thumb & abduction/ adduction & flexion & contact location & manipulability & closure and force \\
\hline \hline
Cylindrical & Yes & No & Irrelevant & Loosely adducted & Resistance-based & Finger and palm & Easy & containment\\
\hline
Funnel & Yes & No &  Yes & Partially abducted & Small & Finger &  Easy & containment\\
\hline
Cup & Yes & No & No & Partially abducted & Large & Finger &  Easy & containment\\
\hline
Tracks & Yes & No & Yes & Partially abducted & Small & Finger &  Easy & containment\\
\hline
Inverse basket & Yes & No & Yes & Partially abducted & Small & Finger and palm & Hard & Force closure\\
\hline
Max & Yes & No & Yes & Fully abducted & Small & Finger and palm & Hard & Force closure\\
\hline
Abduction Clip & No & Yes & Irrelevant & Partially abducted & Irrelevant & Finger side & Hard & Force closure\\
\hline
Finger-palm clip & No & Yes & Irrelevant & Irrelevant & Resistance-based & Finger and palm & Hard & Force closure\\
\hline
Finger-finger clip & No & Yes &  Yes & Fully abducted & Resistance-based & Finger & Hard & Force closuree\\
\hline
Fingertip-fingertip pinch & No & Yes & Yes & Fully abducted & Resistance-based & Fingertip & Hard & Force closure\\
\hline
Fingertip-finger pinch & No & Yes & Yes & Fully abducted & Resistance-based & Finger &  Hard & Force closure\\
\hline
Multi-finger pinch & No & Yes & Yes & Partially abducted & Resistance-based & Fingertip &  Hard & Force closure\\
\hline
\end{tabular}
\caption{Characteristics of each MOG type.}
\label{table-type-groups}
\end{center}
\end{table*}

\subsection{Taxonomy}
\begin{figure}[h]
    \centering
    \vspace{-.cm}
    \includegraphics[width=0.5\textwidth]{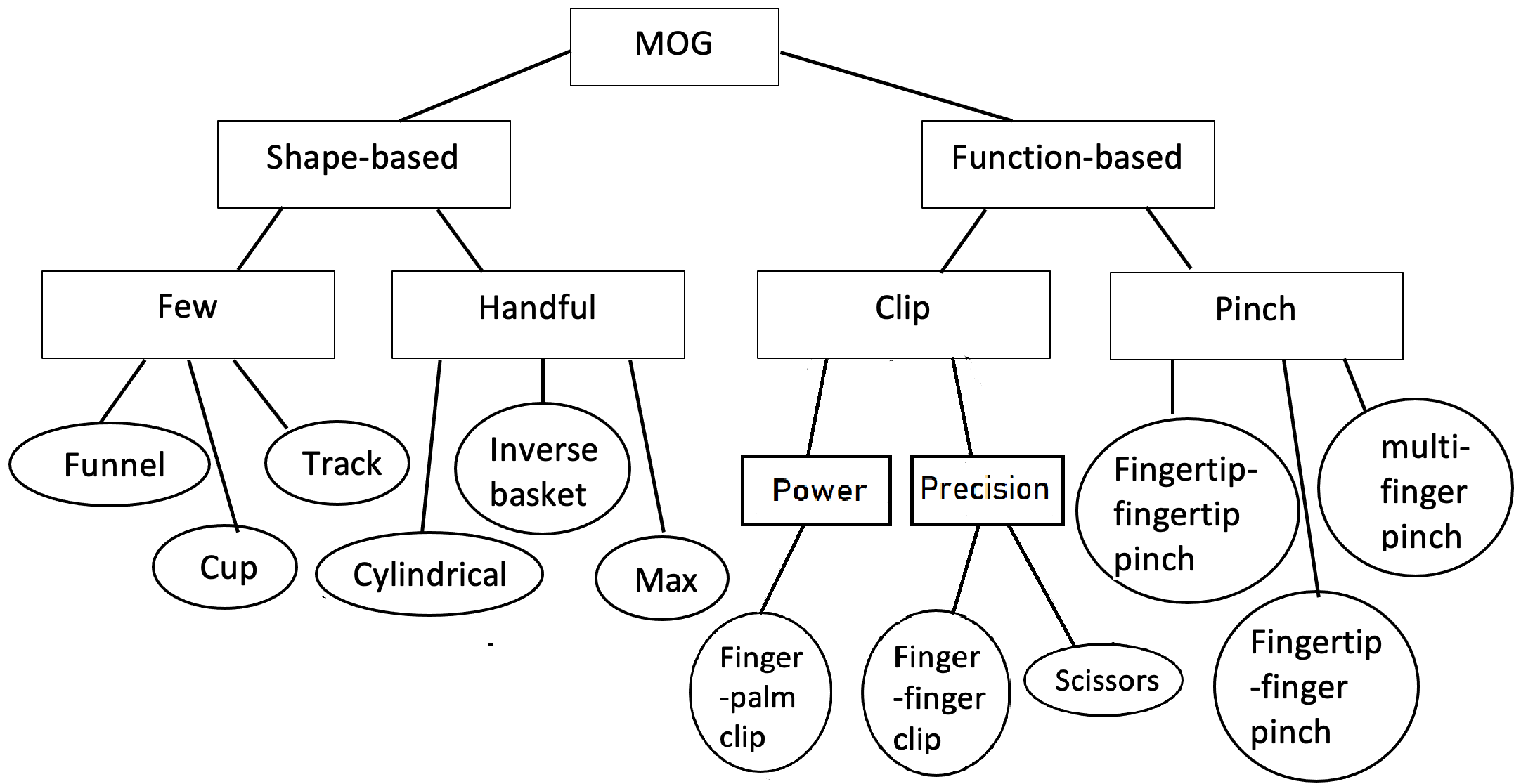}
    \caption{MOG Taxonomy.}
    \vspace{-.cm}
    \label{fig-tax}
\end{figure}

We can organize the MOG types into several taxonomies based on their groups and characteristics. Figure \ref{fig-tax} presents one taxonomy first based on their groups. In the shape-based group, the types can be divided by the quantity of the objects in grasps. While in the function-based group, the types are divided into {\it Clip} and {\it Pinch}. We can further divide the types under {\it Clip} into two groups: using the palm and not using the palm. 

\subsection{Combination of types} 
Different from single-object grasping, multi-object grasping types can be combined to form more complicated grasp types or have multiple same grasp types in one grasp.  The popular duplicated grasp types in one grasp are:  Multiple Scissors shapes;  Multiple finger - palm clips, 
Multiple finger - finger clips;
Multiple fingertip - fingertip pinches;
Multiple lateral pinches;
Multiple Tripods. 

We have observed that many function-based grasp types are regularly combined:
Cylindrical \& fingertip - fingertip pinch; 
Cylindrical \& Scissors;
Funnel \& Scissors;
Funnel \& fingertip-fingertip pinch;
Cup \& Scissors;
Cup \& finger-palm clip;
Tracks \& Scissors;
Tracks \& finger - palm clip;
Inverse basket \& Scissors;
Max \& Scissors;
and any combinations of function-based grasp types. 

Table \ref{table-hybrid} shows several examples of the combined grasp types. 

\begin{table*}[t!]
\begin{center}
\begin{tabular}{ |p{1.8cm}|p {2cm}|p {1.8cm}|p {1.8cm}|p {1.8cm}|p {1.8cm}|p {1.8cm}|p {1.8cm} |} 
\hline
Multiple Scissors & Multiple finger - palm clips &  Multiple finger - finger clips &   Multiple fingertip - fingertip pinches &  Multiple lateral pinches &  Multiple Tripods & Cylindrical \& fingertip - fingertip pinch & Cylindrical \& Scissors \\ \hline
\includegraphics[width=2cm]{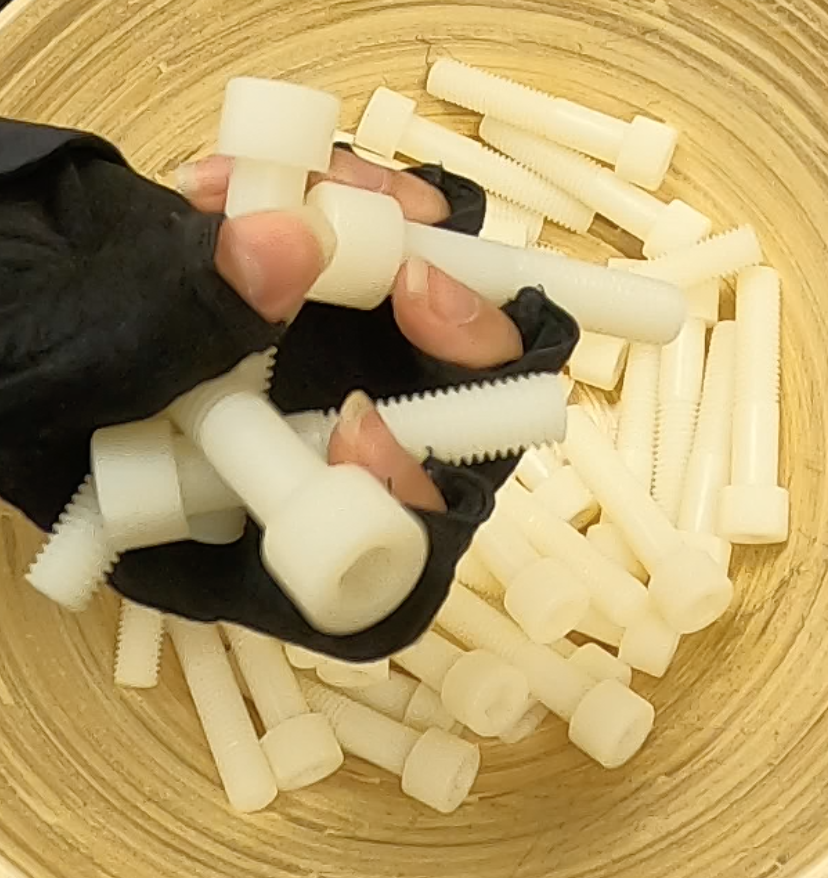} &
\includegraphics[width=2cm]{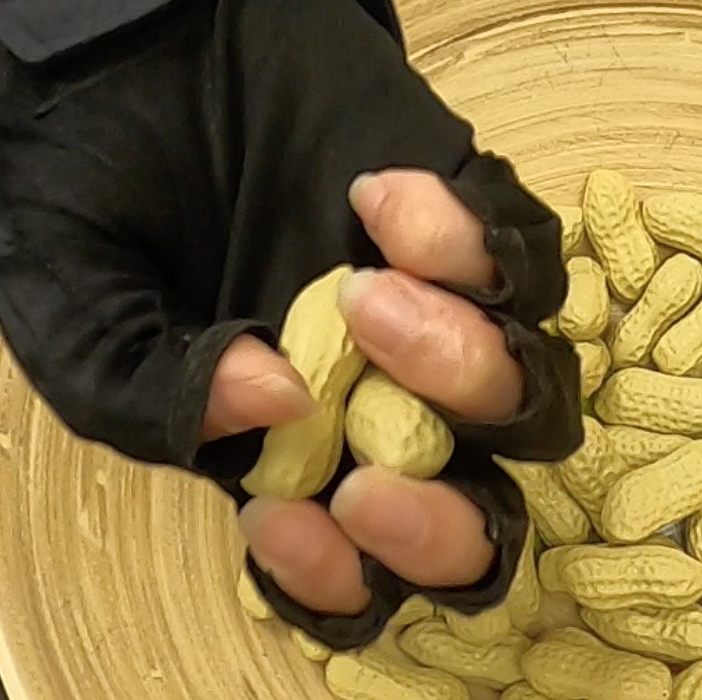} &
\includegraphics[width=2cm]{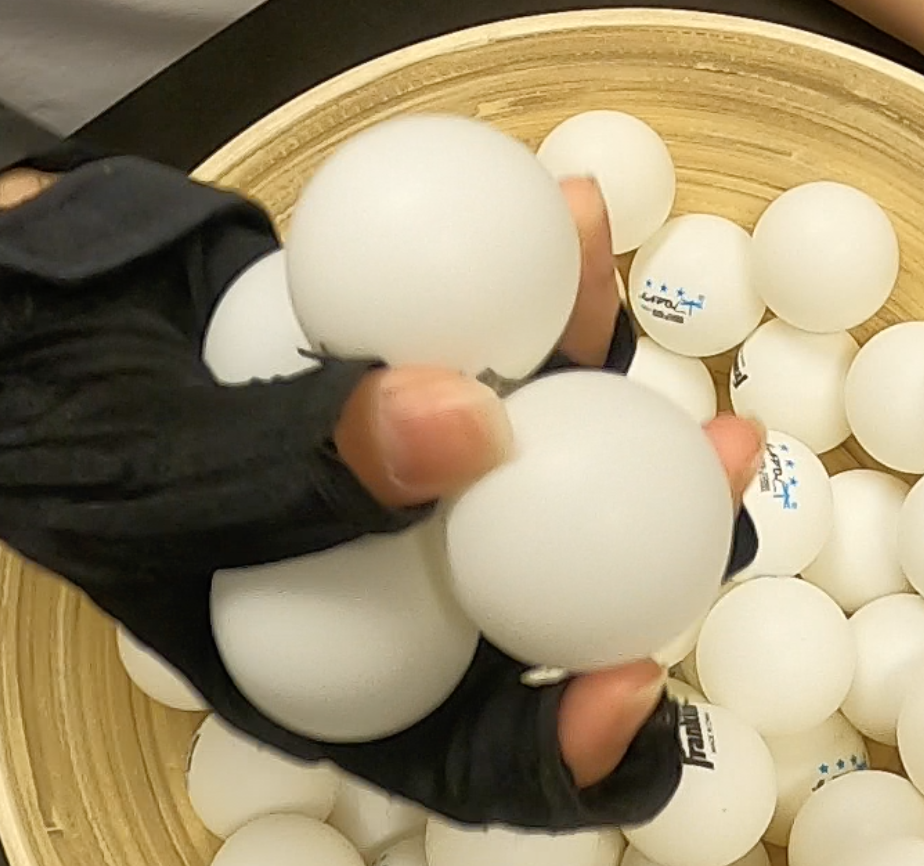} &
\includegraphics[width=2cm]{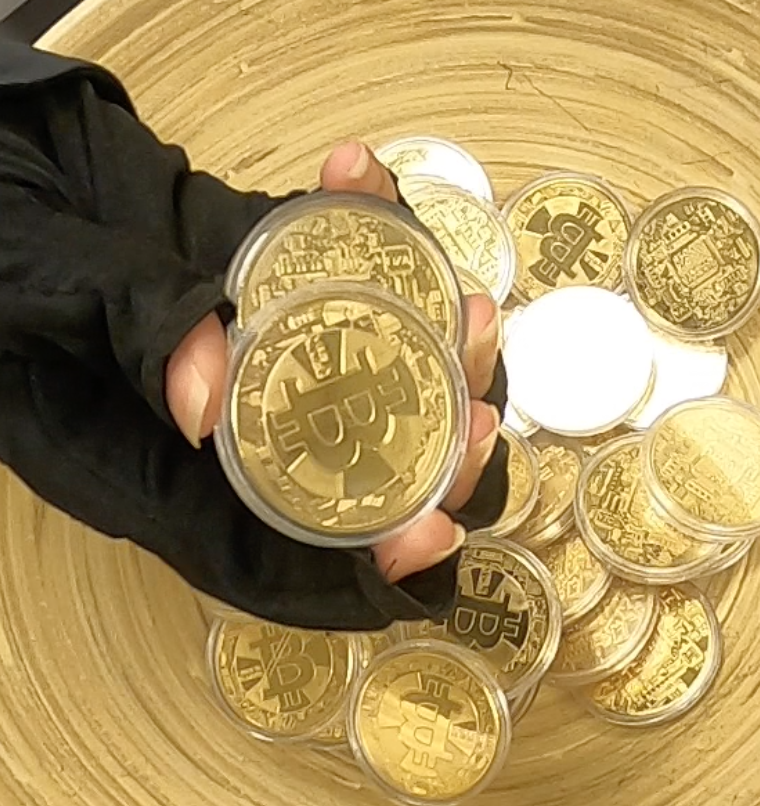} & \includegraphics[width=2cm]{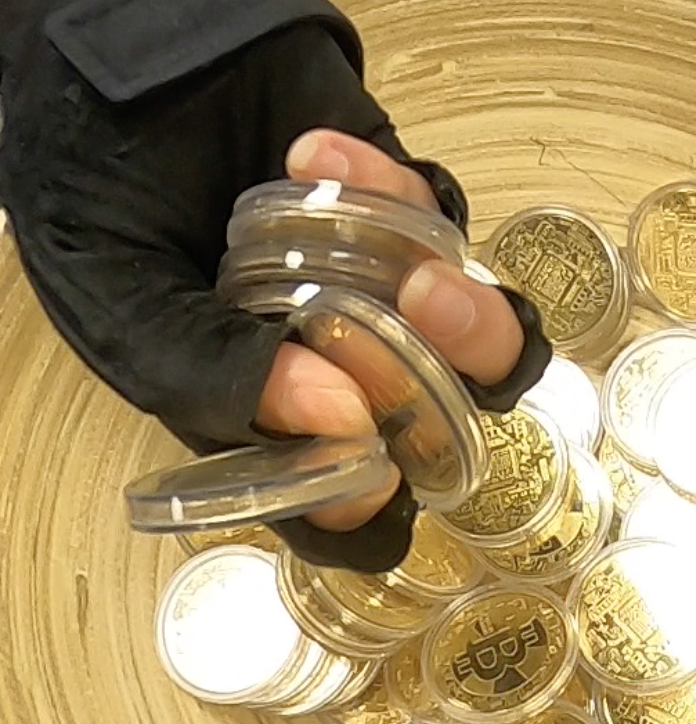} & \includegraphics[width=2cm]{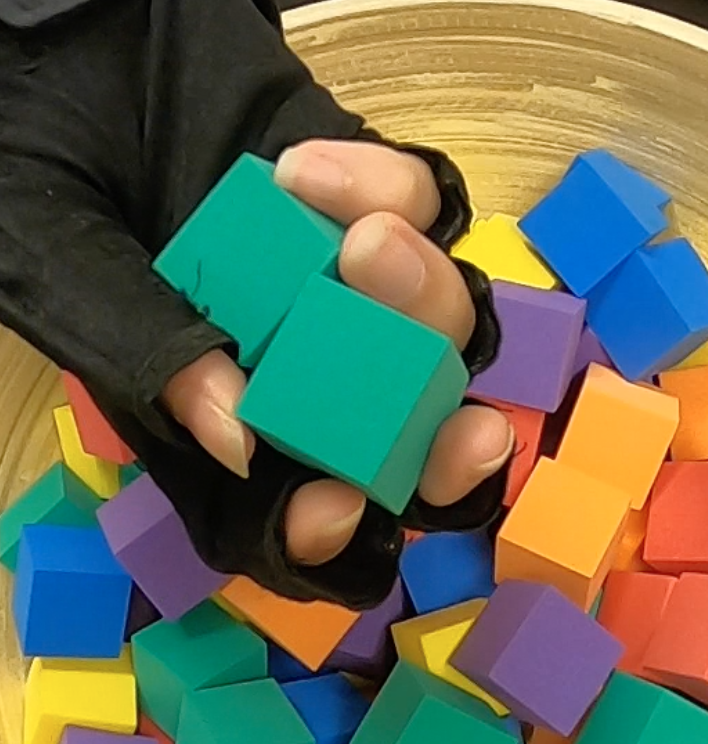} &
\includegraphics[width=2cm]{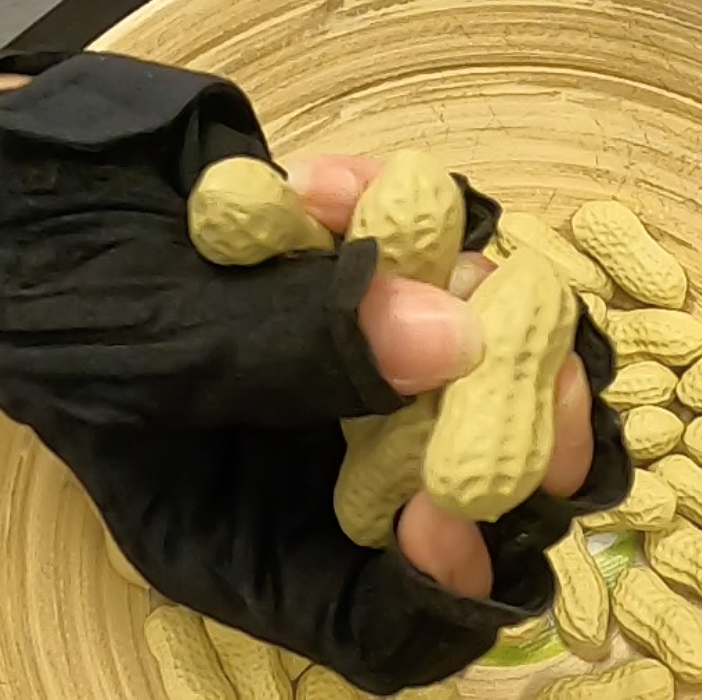} &
\includegraphics[width=2cm]{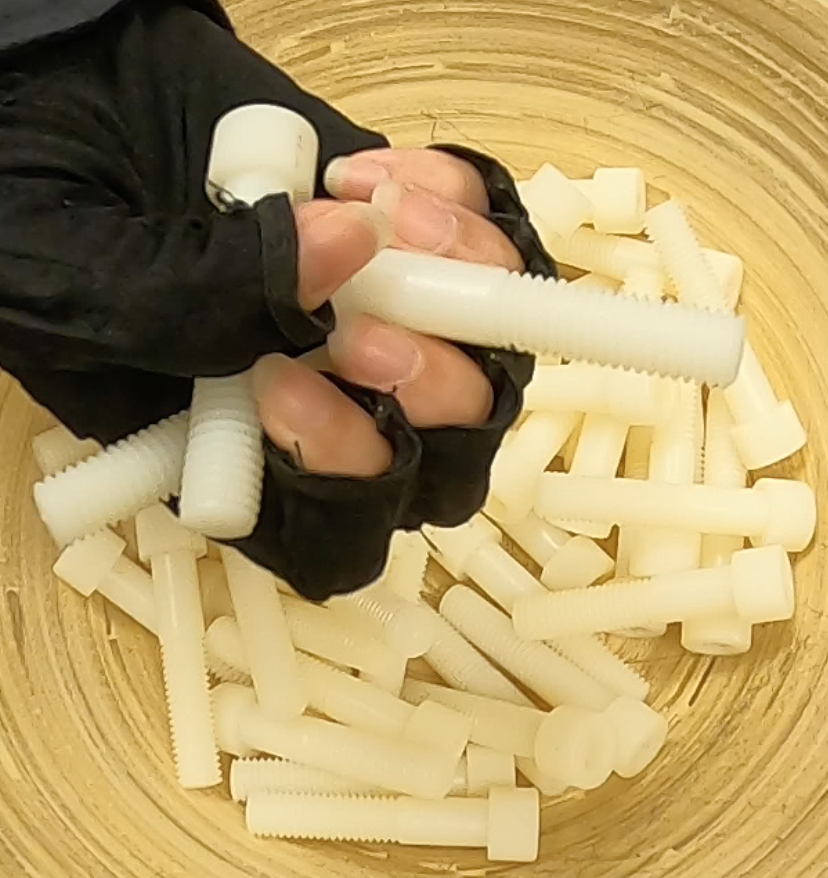} \\ 
\hline
\includegraphics[width=2cm]{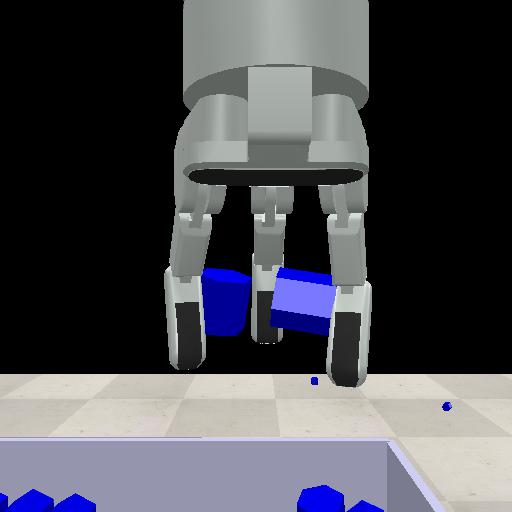} &
\includegraphics[width=2cm]{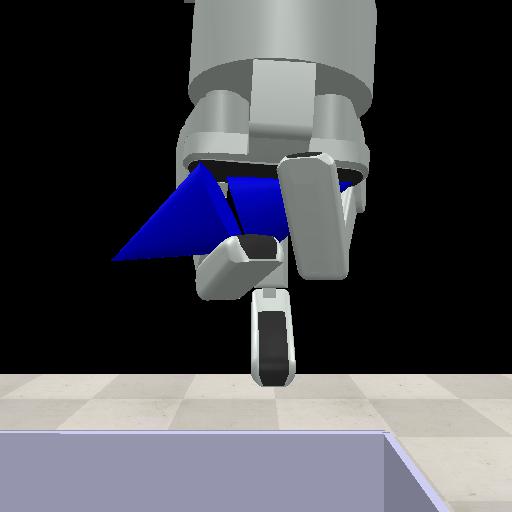} &
N/A &
\includegraphics[width=2cm]{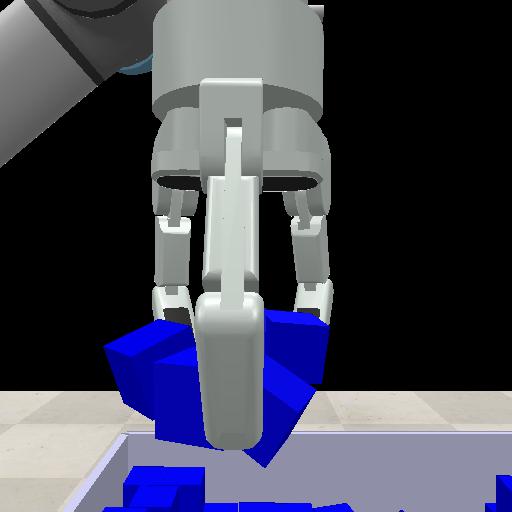} & \includegraphics[width=2cm]{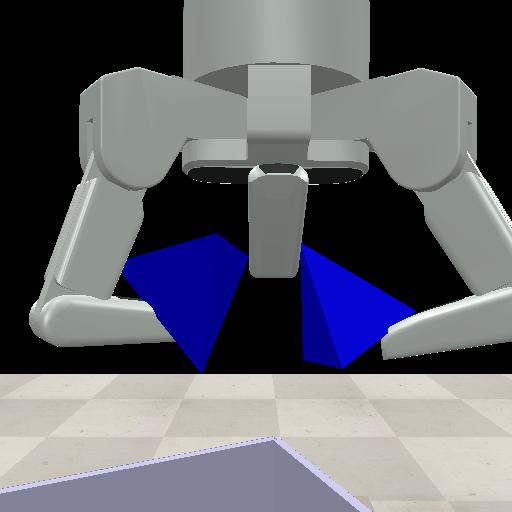} & 
 N/A &
\includegraphics[width=2cm]{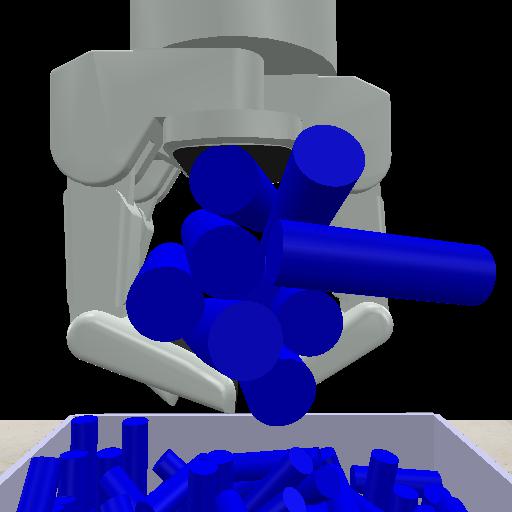} &
\includegraphics[width=2cm]{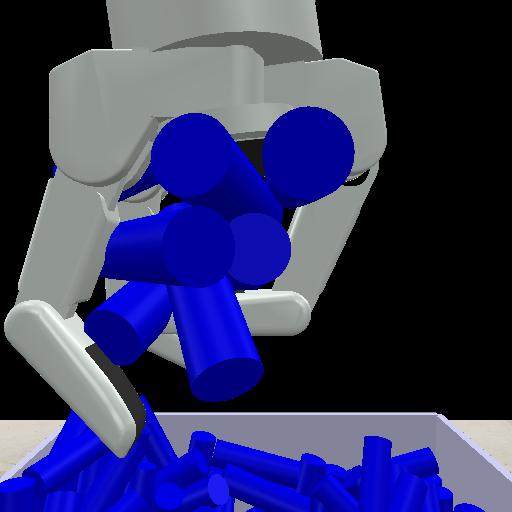} \\ \hline 
\hline
 Funnel \& Scissor& Funnel \& Fingertip-fingertip pinch & Cup \& Scissors & Cup \& Finger-palm clip & Tracks \& Scissors & Tracks \& Finger - palm clip & inverse basket \& Scissors & Max \& Scissors \\ \hline 
\includegraphics[width=2cm]{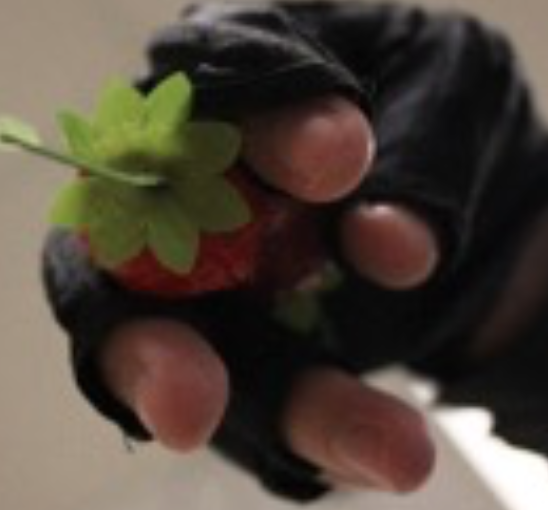} &
\includegraphics[width=2cm]{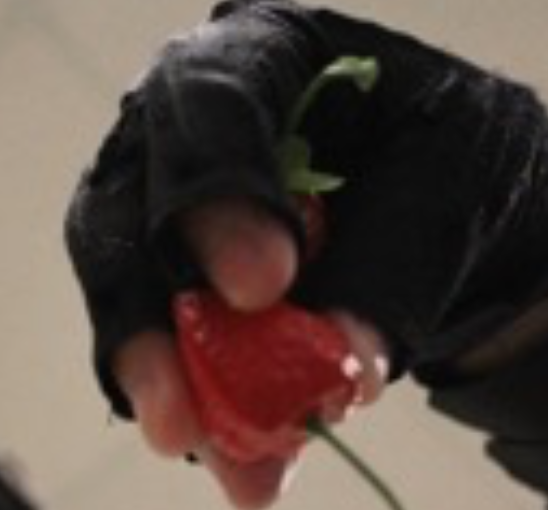} &
\includegraphics[width=2cm]{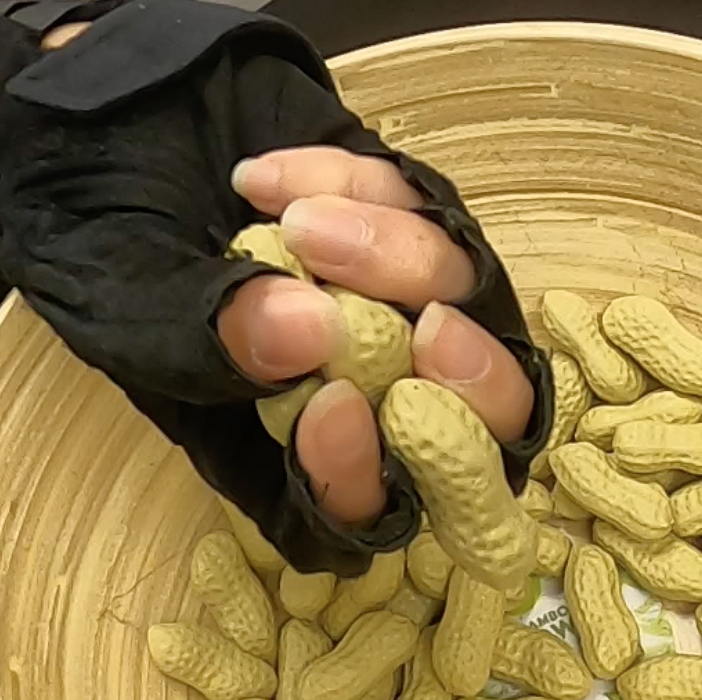} &
\includegraphics[width=2cm]{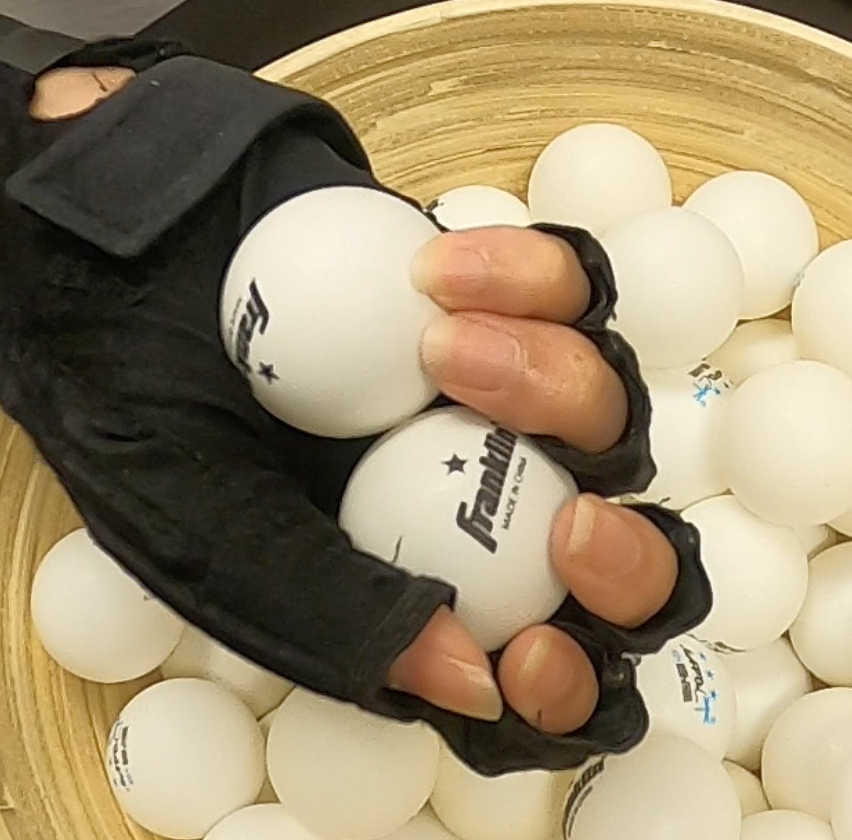} & 
\includegraphics[width=2cm]{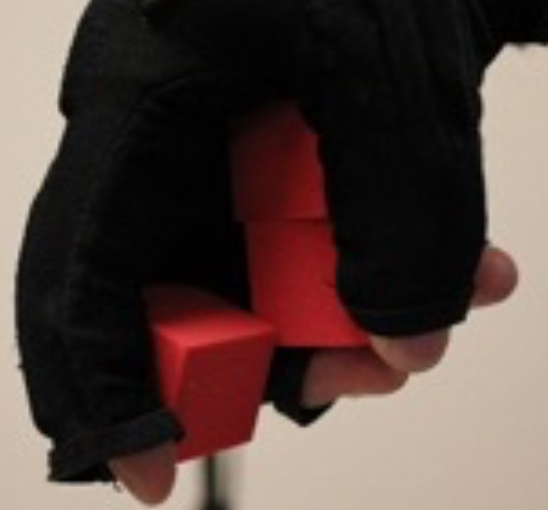} & 
\includegraphics[width=2cm]{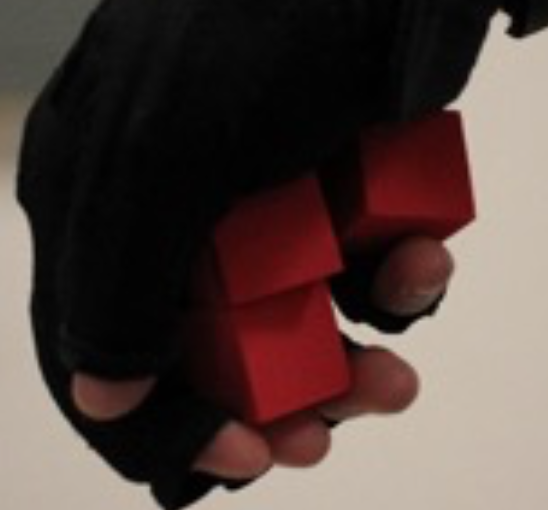} &
\includegraphics[width=2cm]{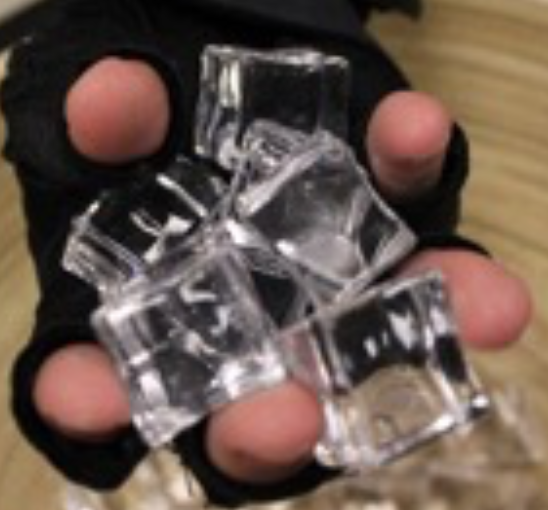} &
\includegraphics[width=2cm]{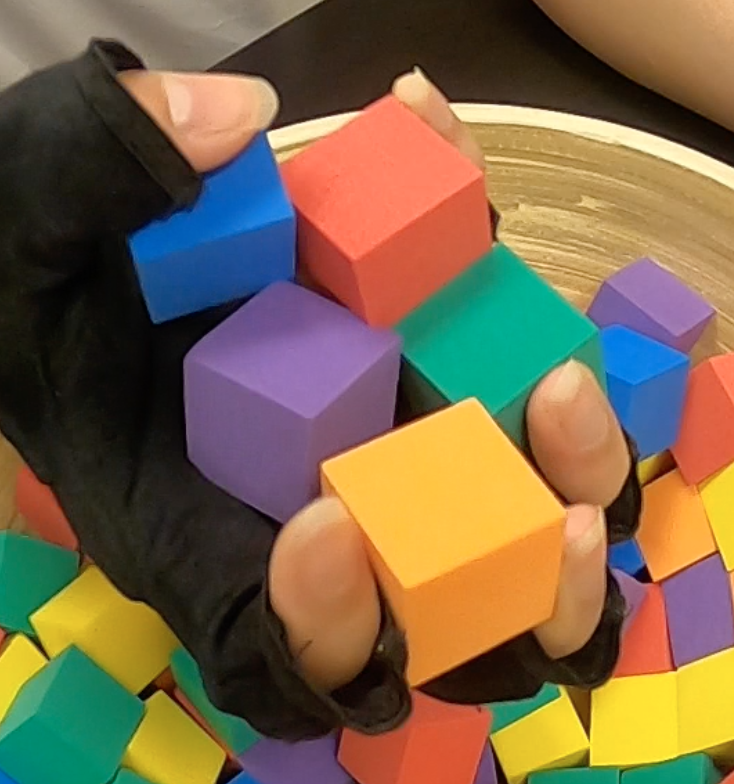} 
 \\ \hline
\includegraphics[width=2cm]{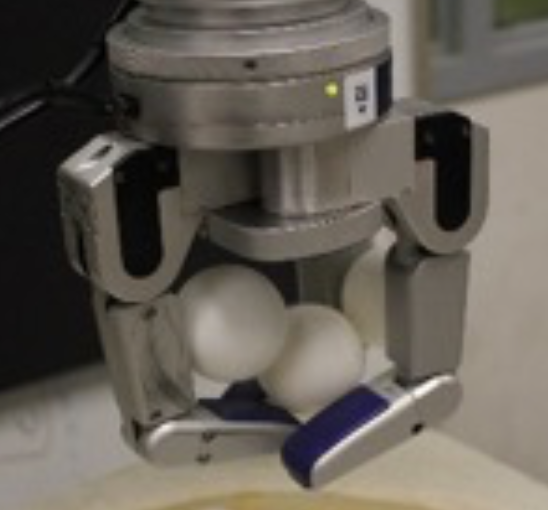} &
 \includegraphics[width=2cm]{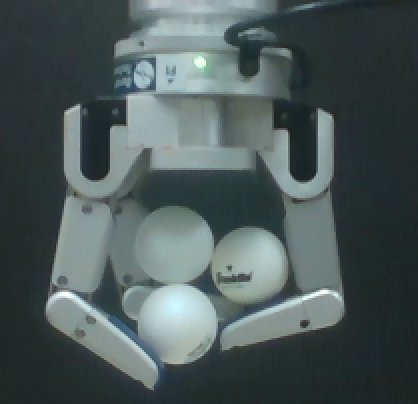} &
 N/A &
\includegraphics[width=2cm]{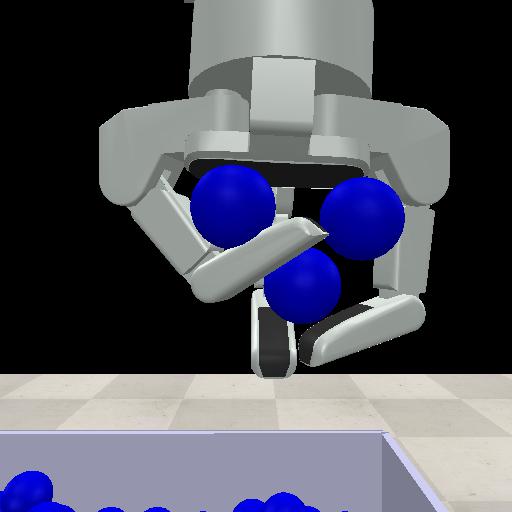} &
N/A & 
\includegraphics[width=2cm]{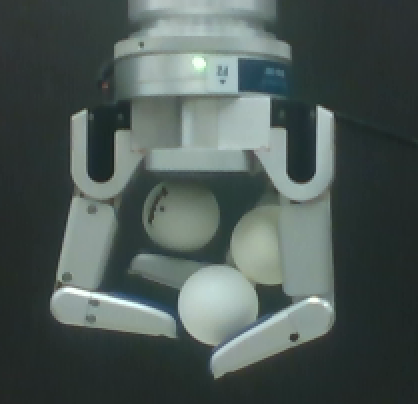} & 
N/A&
N/A
 \\ \hline 

\end{tabular}
\caption{Multiple and hybrid types.}
\label{table-hybrid}
\end{center}
\end{table*}

\section{Conclusion}
We have developed three MOG setups to collect MOG data in human demonstrations, stochastic grasping in a robotic simulation, and stochastic grasping in a real robotic system. We developed a novel stochastic grasping routine based on bias random walk to fully explore the robotic hand's configuration space for feasible MOGs. Using the setups and data collection routines, we have collected 400 human MOGs and 27,800 robotic MOGs. After studying the collected data manually, we have summarized them in 12 MOG types of two groups: shape-based types and function-based types. We then further study the new MOG types using six characteristics based on hand configurations, contact features, and manipulation difficulties. We then compiled the MOG types into the first MOG taxonomy.  

We have also observed that many MOGs could belong to multiple grasp types since some objects are held with several fingers in one type while others are held with the same and/or other fingers in other types. Therefore, we introduce observed MOG type combinations and show examples of 16 different combinations. We have also compared the MOG types with single-object types and taxonomies.  

Similar to SOG types and taxonomy, we believe that the MOG types and taxonomy are useful in developing suitable MOG planning algorithms, analyzing multi-object containment, and perceiving objects status in MOGs. This study is limited to one robotic hand and one human hand because the authors' lab has only a Barrett Hand. Therefore, the MOG types discovered in this study are only common types. Studies with other robotic hands may produce different MOG types and taxonomy. It is also likely people with extraordinary skills can produce more diverse MOG types. Even though this study does not explore the relationship between MOG and in-hand manipulations \cite{harada2000rolling}, general manipulations \cite{lin2015grasp, lin2016task, lin2015task}, or novel robotic gripper/hand designs \cite{mucchiani2020novel}, the MOG types could be helpful in those research directions.

\section*{Acknowledgment}
This material is based upon work supported by the National Science Foundation under Grants Nos. 1812933 and 191004.

\bibliographystyle{IEEEtran}

\bibliography{references, sun}

\end{document}